\documentclass[a4paper,twoside]{article}

\usepackage{amsmath}
\usepackage{amsthm}
\usepackage{fullpage}
\usepackage{physics}
\usepackage[linesnumbered,ruled,vlined]{algorithm2e}
\usepackage{color}
\usepackage{url}
\usepackage{epsfig}
\usepackage{subcaption}
\usepackage{calc}
\usepackage{amssymb}
\usepackage{amstext}
\usepackage{multicol}
\usepackage{pslatex}
\usepackage{apalike}
\usepackage{listings}
\usepackage[bottom]{footmisc}
\usepackage{booktabs}
\usepackage{multirow}
\usepackage{caption}
\usepackage{graphicx}
\usepackage{lipsum}
\usepackage[table,xcdraw]{xcolor}
\usepackage{adjustbox}
\usepackage{graphicx}
\usepackage{SCITEPRESS}  
\begin{document}

\twocolumn

\title{Comparative Evaluation of Metaheuristic Algorithms for Hyperparameter Selection in Short-Term Weather Forecasting}

\author{\authorname{Anuvab Sen\sup{1}\orcidAuthor{0009-0001-8688-8287}, Arul Rhik Mazumder\sup{2}\orcidAuthor{0000-0002-2395-4400}, Dibyarup Dutta\sup{3}\orcidAuthor{0009-0008-9012-2816}, Udayon Sen\sup{4}\orcidAuthor{0009-0006-6575-6759}, Pathikrit Syam\sup{1}\orcidAuthor{0009-0003-2370-182X}, and Sandipan Dhar\sup{5}\orcidAuthor{0000-0002-3606-6664}}
\affiliation{\sup{1}Electronics and Telecommunication, Indian Institute of Engineering
Science and Technology, Shibpur, Howrah, India}
\affiliation{\sup{2}School of Computer Science, Carnegie Mellon University, Pittsburgh, Pennsylvania, United States of America}
\affiliation{\sup{3}Information Technology, Indian Institute of Engineering
Science and Technology, Shibpur, Howrah, India}
\affiliation{\sup{4}Computer Science and Technology, Indian Institute of Engineering
Science and Technology, Shibpur, Howrah, India}
\affiliation{\sup{5}Computer Science and Engineering, National Institute of Technology, Durgapur, West Bengal, India}
\email{sen.anuvab@gmail.com, arul.rhik@gmail.com, dibyarupdutta@gmail.com, udayon.sen@gmail.com, pathikritsyam@gmail.com, sd.19cs1101@phd.nitdgp.ac.in}
}


% \abstract{The abstract should summarize the contents of the paper and should contain at least 70 and at most 200 words. The text must be set to 9-point font size.}

\abstract{Weather forecasting plays a vital role in numerous sectors, but accurately capturing the complex dynamics of weather systems remains a challenge for traditional statistical models. Apart from Auto Regressive time forecasting models like ARIMA, deep learning techniques (Vanilla ANNs, LSTM and GRU networks), have shown promise in improving forecasting accuracy by capturing temporal dependencies. This paper explores the application of metaheuristic algorithms, namely Genetic Algorithm (GA), Differential Evolution (DE), and Particle Swarm Optimization (PSO), to automate the search for optimal hyperparameters in these model architectures. Metaheuristic algorithms excel in global optimization, offering robustness, versatility, and scalability in handling non-linear problems. We present a comparative analysis of different model architectures integrated with metaheuristic optimization, evaluating their performance in weather forecasting based on metrics such as Mean Squared Error (MSE) and Mean Absolute Percentage Error (MAPE). The results demonstrate the potential of metaheuristic algorithms in enhancing weather forecasting accuracy \& helps in determining the optimal set of hyper-parameters for each model. The paper underscores the importance of harnessing advanced optimization techniques to select the most suitable metaheuristic algorithm for the given weather forecasting task.}

\onecolumn \maketitle \normalsize \setcounter{footnote}{0} \vfill

% \section{\uppercase{Introduction}}
% \label{sec:introduction}
% Genetic Algorithm (GA) has been prevalent for more than six decades which first came into existence as a result of the research of John Holland in the University of Michigan around 1960. But they remained unpopular until late 20\textsuperscript{th} Century. It follows the Darwinian theory of Natural Selection which is based upon a bigger class of Evolutionary Algorithms. GA is widely used to create solutions to search problems by bio-inspired operatives like mutation, recombination, and selection processes.

% Your paper will be part of the conference proceedings
% therefore we ask that authors follow the guidelines explained in
% this example in order to achieve the highest quality possible

% \cite{Smith98}.

% Be advised that papers in a technically unsuitable form will be
% returned for retyping. After returned the manuscript must be
% appropriately modified.

\section{\uppercase{Introduction}}
Weather forecasting is the use of science and technology to predict future weather conditions in a specific geographical area. It plays a vital role in agriculture, transportation, and disaster management. Traditional methods rely on physical models, but they may struggle to capture the complex dynamics of weather systems accurately. To address this, deep learning techniques have emerged, leveraging large datasets to improve forecasting accuracy by uncovering hidden patterns.
\par

Recently, time series models like recurrent neural networks (RNNs) have become popular in weather forecasting due to their ability to capture temporal dependencies. However, the standard RNN model often faces challenges like exploding and vanishing gradient problems, making it difficult to capture long-term dependencies. To address this, Long Short-Term Memory (LSTM) models have emerged as superior alternatives \cite{Hochreiter_Schmidhuber_1997}, excelling at capturing sequential information from long-term dependencies. Additionally, Gated Recurrent Unit (GRU) networks \cite{Jing_Gulcehre_Peurifoy_Shen_Tegmark_Soljacic_Bengio_2019}, another class of RNNs, have shown promise in sequence prediction problems. GRUs mitigate the vanishing gradient problem by employing update and reset gates, significantly improving the modeling of long-term dependencies. Moreover, due to their reduced parameter count, GRUs generally require less training time than LSTMs.

Achieving optimal performance with GRUs often requires manual tuning of hyperparameters such as learning rate, batch size, and number of epochs, which is time-consuming and labor-intensive. To overcome this, the paper proposes the use of metaheuristic algorithms like Genetic Algorithm (GA), Differential Evolution (DE), and Particle Swarm Optimization (PSO) to automate the search for the best hyperparameter settings, ultimately enhancing forecasting accuracy. These metaheuristic algorithms offer advantages in tackling challenges related to training and optimizing complex neural network architectures.
\par

Weather forecasting saw a notable improvement with the introduction of metaheuristic algorithms to optimize deep learning models like GRU. Leveraging the global optimization capabilities of these algorithms, weather forecasting models achieved better performance by efficiently exploring and exploiting the search space to find optimal solutions. This is essential for accurately predicting complex and dynamic weather systems. Moreover, metaheuristic algorithms exhibit robustness, versatility, and scalability, enabling them to handle non-linear and non-convex problems effectively. Integrating them with existing models facilitates adaptation to evolving challenges in weather prediction.\par

Genetic Algorithm (GA) \cite{Man_Tang_Kwong_1996}, based on the Darwinian theory of Natural Selection, was developed in the 1960s but gained popularity in the late 20th Century. It falls under the broader class of Evolutionary Algorithms and is widely used to solve search problems through bio-inspired processes like mutation, recombination, and selection. Differential Evolution (DE) \cite{Storn_Price_1997} is another population-based metaheuristic algorithm that iteratively improves a population of candidate solutions to optimize a problem. Particle Swarm Optimization (PSO) \cite{Kennedy_Eberhart_1995}, inspired by bird flocking or fish schooling behavior, maintains a swarm of particles that move in the search space to find the optimal solution.\par

This paper presents a comparative analysis of architectures for weather forecasting using a meta-heuristic optimization algorithm. We evaluate these architectures based on metrics like Mean Squared Error (MSE) and Mean Absolute Percentage Error (MAPE). Additionally, we curate a comprehensive weather dataset spanning 10 years to train our best forecasting model. Leveraging the Gated Recurrent Unit (GRU) architecture with Differential Evolution optimization, we achieve superior accuracy and performance in predicting weather conditions.
\section{\uppercase{Related Work}}

Hyperparameter optimization is a critical research area for achieving high-performance models. Techniques like Random Search \cite{Radzi_Karim_Saripan_Rahman_Isa_Ibahim_2021}, Grid Search \cite{8882943}, Bayesian Optimization \cite{Masum_Shahriar_Haddad_Faruk_Valero_Khan_Rahman_Adnan_Cuzzocrea_Wu_2021}, and Gradient-based Optimization \cite{maclaurin2015gradientbased} are used to find optimal hyperparameter configurations. Each method offers trade-offs in computational efficiency, exploration of search space, and exploitation of solutions.

Genetic Algorithms were first utilized for modifying Artificial Neural Network architectures in 1993 \cite{Bäck_Schwefel_1993}, inspiring various nature-based algorithms' applications to deep-learning models \cite{Katoch_Chauhan_Kumar_2020}. While many works compare evolutionary algorithms on computational models, no previous study comprehensively compares the three most promising evolutionary algorithms: Genetic Algorithm, Differential Evolution, and Particle Swarm Optimization, across multiple computational architectures. These algorithms stand out due to their iterative population-based approaches, stochastic and global search implementation, and versatility in optimizing various problems. Moreover, no research has explored these metaheuristics across such a diverse range of models. Although many papers analyze metaheuristic hyperparameter tuning on Artificial Neural Networks \cite{article}, \cite{NEMATZADEH2022107619} and Long Short-Term Memory Models \cite{Wang_Ma_Xu_Wang_Wu_2022}, they are few metaheuristic hyperparameter tuning methods for Auto-Regressive Integrated Moving Average and Gated Recurrent Networks. Additionally, a thorough investigation of papers on metaheuristic applications for ARIMA and GRU tuning reveals that none of them utilize the three evolutionary algorithms (GA, DE, PSO) discussed in this study.
\section{\uppercase{Background}}
\subsection{\uppercase{Metaheuristics}}
\subsubsection{Genetic Algorithm}

Genetic Algorithm (GA) is a metaheuristic algorithm based on the evolutionary process of natural selection, the key driver of biological evolution mentioned in Darwin’s theory of evolution. Similar to how evolution generates successful individuals in a population, GA generates optimal solutions to constrained and unconstrained optimization problems. The summarized pseudocode of the Genetic Algorithm is shown in Algorithm 1.

\begin{algorithm}[htbp]
\SetAlgoLined
\SetKwInOut{Input}{Input}
\SetKwInOut{Output}{Output}
\Input{Population Size $N \in (5, 10)$, Chromosome Length $L$, Termination Criterion}
\Output{Best Individual}

Initialize the population with $N$ random individuals;

\While{Termination Criterion is not met}{
Evaluate fitness $f(x_{i})$ for all individual $x_i$ in population\;

Select parents $p_{1}, p_{2}$ from the population for mating. Use a Roulette Wheel Selection scheme.\;

Create a new population by applying uniform crossover:
and mutation operations on the selected parents to get $c_{i}$\;

Replace the current population with the new population\;
}

\Return best individual in the final population;

\caption{Genetic Algorithm}
\end{algorithm}

Genetic Algorithm initializes a population, selects parents for mating using a custom fitness function, and produces new candidate solutions by applying mutations and crossover to previous solutions. We utilized a Roulette Wheel Selection scheme and a Uniform Crossover scheme and ran Genetic Algorithm for 10 generations. 

\subsubsection{Differential Evolution}

Differential Evolution (DE) is a population-based metaheuristic used to solve non-differentiable and non-linear optimizations. DE obtains an optimal solution by maintaining a population of candidate solutions and iteratively improving these solutions by applying genetic operators.

Similar to the Genetic Algorithm, Differential Evolution begins
by randomly initializing a population and then generating  new solutions using crossover and mutation operators, shown below. It uses a custom fitness function when deciding to replace previous solutions. 
\vspace{0.2cm}

A pseudocode implementation for Differential Evolution is provided in Algorithm 2. \noindent We cycled through the selection, mutation, and crossover operations of Differential Evolution for 10 generations be-

\begin{algorithm}[htbp]
\SetAlgoLined
\SetKwInOut{Input}{Input}
\SetKwInOut{Output}{Output}
\Input{Population Size $N \in (5, 10)$, Dimension $D$, Scale Factor $F$, Crossover Probability $CR$, Termination Criterion}
\Output{Best individual}

Initialize the population with $N$ random individuals in the search space;

\While{Termination Criterion is not met}{
\For{each individual $x_i$ in the population}{
Select three distinct individuals $x_{r_{1}}$, $x_{r_{2}}$, and $x_{r_{3}}$ from the population;

Generate a trial vector $v_i$ by mutating $x_{r_{1}}$, $x_{r_{2}}$, and $x_{r_{3}}$ using the differential weight $F$;
\begin{equation}
    v_{i} = x_{r_{1}} + F \times (x_{r_{2}}-x_{r_{3}})
\end{equation}

Perform crossover between $x_i$ and $v_i$ to produce a trial individual $u_i$ with the crossover probability $CR$;
\begin{equation}
u_{j, i} = 
\left\{
    \begin{array}{lr}
        v_{j, i}, & \text{if } p_{rand}(0, 1) \leq CR \\
        x_{j, i} & \text{else} 
    \end{array}
\right\}
\end{equation}

    \If{the fitness of $u_i$ is better than the fitness of $x_i$}{
        Replace $x_i$ with $u_i$ in the population\;
    }
}
}

\Return the best individual in the final population;

\caption{Differential Evolution}
\end{algorithm}
\noindent fore terminating the number of runs.

\subsubsection{Particle Swarm Optimisation}
Particle Swarm Optimization (PSO) is a population-based metaheuristic algorithm inspired by the social behavior of bird flocking or fish schooling. It aims to find optimal solutions by simulating the movement and interaction of particles in a multidimensional search space. Like other metaheuristic algorithms, PSO begins with the initialization of particles and arbitrarily sets their position and velocity. Each particle represents a potential solution, and their positions and velocities are updated iteratively based on a fitness function and the global best solution found by the swarm. The summarized pseudocode of PSO is displayed in Algorithm 3.

\begin{algorithm}[htbp]
\SetAlgoLined
\SetKwInOut{Input}{Input}
\SetKwInOut{Output}{Output}
\Input{Number of particles $N \in (5, 10)$, Max Iterations $M$, Termination Criterion}
\Output{Global Best Fitness}
Initialize the particle's position with a uniformly distributed random vector: $x_i \sim U(b_{lo}, b_{up})$;

Initialize the particle's velocity: $v_i \sim U(-|b_{up}-b_{lo}|, |b_{up}-b_{lo}|)$;

Calculate Global Best Fitness $f(g)$;

\For{$i = 1$ \KwTo $M$}{
    \For{$j = 1$ \KwTo $N$}{
    Update the particle's position: $x_i \gets x_i + v_i$;
    
    Update the particle's velocity: $v_{i} \gets \omega v_{i} + c_1 r_1 (p_{i}-x_{i}) + c_2 r_2 (p_{g}-x_{i})$;
    
    Evaluate fitness $f(p_i)$;
    
    \If{$f(p_i) < f(g)$}{
         $f(g) = f(p_i)$
         
        $x_n = x_{n+1}$
    }
    }
}
\caption{Particle Swarm Optimization}
\end{algorithm}

%\paragraph{Algorithm}
%Let \( f: \mathbb{R}^n \to \mathbb{R} \) be the cost function that must be minimized. In this work, we used the mean squared error (MSE) loss function metric for our cost function. The function takes a candidate solution as an argument in the form of a vector of real numbers and produces a real number as output, indicating the objective function value of the given candidate solution. The goal is to find a solution \( a \) for which \( f(a) \leq f(b) \) for all \( b \) in the search space, where \( a \) is the global minimum.

\noindent  In Algorithm 3, $b_{lo}$ \& $b_{up}$ are bounds on the range of values within which the initialization parameters will be initialized.

\subsection{\uppercase{Models}}
\subsubsection{Auto-Regressive Integrated Moving Average}
The Auto-Regressive Integrated Moving Average (ARIMA) \cite{Harvey_1990} is a time series forecasting model that extends the autoregressive moving average (ARMA) model \cite{Brockwell_Davis_1996}. Their advantage, is that they can reduce a non-stationary series into a stationary series and thus provide a broader scope of forecasting capabilities. An ARIMA model is composed of three components:
\begin{enumerate}
    \item Auto-regression (AR) measures a variable's dependence on its past values, enabling predictions of future values. In ARIMA, the $p$ hyperparameter represents this order and the number of past values used for current observations. A time series $\{x_{i}\}$ is said to be autoregressive of order $p$ if:
    \begin{equation}
        x_{i} = \Sigma_{j=1}^{p}\alpha_{j}x_{i-j}+w_{i}
    \end{equation}
    where $w_{i}$ is the white noise and $\alpha_{i}$ are non-zero constant real coefficients.
    \item Integration (I) assimilates time series data, transforming non-stationary series into stationary ones through differencing current and previous values. The hyperparameter $d$ denotes the magnitude of differencing needed for data stationarity.
    \item Moving Average (MA) accounts for past error residuals' impact on the variable value. Like the AR component, it predicts current values using past error combinations. The $q$ hyperparameter determines the number of lagged error terms for the prediction. A time series has a moving average of order $q$ if:
    \begin{equation}
        x_{i} = w_{i} + \beta_{1}w_{i-1} + .... + \beta_{q}w_{i-q}
    \end{equation}
    where $\{w_{i}\}$ is the set of white noise and $\beta_{i}$ are non-zero constant real coefficients.
\end{enumerate}

\subsubsection{Artificial Neural Network}
Artificial Neural Network (ANN) is a computational model inspired by the human brain's neural circuits. A core part of Deep Learning, ANNs detect patterns, learn from historical data, and make informed decisions.

% ANN is a powerful computer architecture due to its ability to restructure its internal structure to provide optimal solutions and provided enough data. ANNs are classified into two categories: supervised and unsupervised, based on their learning processes. This work only utilized supervised ANNs and their implementation is outlined below: 

Neural networks consist of artificial neurons arranged in a multilayer graph. There are three types of layers: input, hidden, and output. Input data is processed through at least one hidden layer, and each neuron in the hidden layer learns patterns from inputs $x_i$ (size $n$) and produces output $h$ using Equation $5$.

\begin{equation}
    h = \rho(b_j + \sum_{i=1}^{n}w_{i}x_{i})
\end{equation}

In the equation above $b_j$ and $w_i$ represent the biases and weights from each input node respectively. The function $\rho$ is the activation function designed to introduce nonlinearity and bound output values. 

Training the neural network follows supervised learning, where biases $b_j$ and weights $w_i$ are adjusted to achieve the optimal output. During training, an error function $f$ compares the ANN's output with the desired output for each data point in the training set. The errors are then corrected through Backpropagation \cite{Kelley_1960}, a stochastic gradient descent algorithm. The initial weights are adjusted according to Equations $6$ and $7$ below:
\vspace{0.20cm}
\begin{equation}
 w := w - \epsilon_{1} \pdv{f}{w}
\end{equation}
\begin{equation}
 b := b - \epsilon_{2} \pdv{f}{b}
\end{equation}
% In the equation above $\epsilon_{1}$ and $\epsilon_{2}$ represent the learning rate and influence how quickly the optimal weights and biases are attained. Backpropagation is executed until the error function is constrained under the present bound at which point the model will be sufficiently trained.
\subsubsection{Long Short-Term Memory}
Long Short-Term Memory (LSTM) is a type of Recurrent Neural Network (RNN) that maintains short-term memory over time by preserving activation patterns. Unlike standard feed-forward neural networks, LSTM's feedback connections allow it to handle data sequences, making it ideal for time series analysis \cite{Hochreiter_Schmidhuber_1997}.

LSTM utilizes specialized memory cells that retain activation patterns across iterations. Each LSTM memory cell comprises four components: a cell, an input gate, a forget gate, and an output gate. The cell serves as the memory for the network. It retains essential information throughout the processing of the sequence. The Input Gate updates the cell. This process is done by passing the previously hidden layer ($h_{t-1}$) information (weights $w_{i}$ and biases $b_{i}$) and current state $x_t$ into a sigmoid function $\sigma$ as outlined in the Equation $8$.
\begin{equation}
    i = \sigma(w_i[h_{t-1}, x_t]+b_i)
\end{equation}
The Forget Gate is responsible for deciding which information is thrown away or retained. This function is very similar to the Input Gate and outlined in Equation $9$ below.
\begin{equation}
    f = \sigma(w_f[h_{t-1}, x_t]+b_f)
\end{equation}
The Output Gate is responsible for determining the next hidden state and is important for predictions. The output function follows from the same functions described with the Input Gate and Forget Gate.
\begin{equation}
    o = \sigma(w_o[h_{t-1}, x_t]+b_o)
\end{equation}

\subsubsection{Gated Recurrent Unit Networks}
The Gated Recurrent Unit (GRU) is an RNN architecture that overcomes some limitations of LSTM while delivering comparable performance. GRUs excel at capturing long-term dependencies in sequential data.

GRUs are computationally less expensive and easier to train than LSTMs due to their simpler architecture. They merge the cell and hidden state, removing the need for a separate memory unit. GRUs also use gating mechanisms to control information flow within the network. The GRU relies on a series of gates similar to the LSTM.
The Update gate determines how much of the previous hidden state to keep and how much of the new input to incorporate. It is computed as:
\begin{equation}
   z_t=\sigma(W_z\cdot\lbrack h_{t-1},x_t\rbrack)
\end{equation}
where $W_z$ is the weight matrix associated with the update gate, $h_t-1$ is the previous-hidden state, $x_t$ is the input at time step $t$ and $\sigma$ is the sigmoid activation function. The Reset Gate determines how much of the previous hidden state to forget. It is computed as:
\begin{equation}
r_t = \sigma(W_r \cdot [h_{t-1}, x_t])
\end{equation}
where $W_r$ is the weight matrix associated with the reset gate. The smaller the value of the reset gate $r_t$ is, the more the state information is ignored. The Current Memory Content is calculated as a combination of the previous hidden state and the new input, controlled by the update gate:
\begin{equation}
\tilde{h}_t = \tanh(W \cdot [r_t \odot h_{t-1}, x_t])
\end{equation}
where $W$ is the weight matrix and $\odot$ denotes element-wise multiplication. The Hidden State at time step $t$ is updated by considering the current memory content.
\begin{equation}
h_t = (1 - z_t) \odot h_{t-1} + z_t \odot \tilde{h}_t
\end{equation}
where $\odot$ denotes element-wise multiplication.
\section{\uppercase{Dataset Description}}
We have created a dataset \footnote{Dataset link: \url{https://doi.org/10.7910/DVN/PJISJU}} by scraping from the official website of Government of Canada \cite{Canada_2023}, by taking weather related data for the region of Ottawa from date 1\textsuperscript{st} January 2010 to 31\textsuperscript{st} December 2020. It has the following features: date, time (in 24 hours), temperature (in $^\circ$ Celsius), dew point temperature (in $^\circ$ Celsius), relative humidity (in $\%$), wind speed (in kilometers/hours), visibility (in kilometers), pressure (in kilopascals) and precipitation amounts (in millimeters). It also had a few derived features like humidity index and wind chill, which we did not take into account to keep our list of features as independent as possible from each other. The compiled data set comprised 96,408 rows of data for 8 variables, where each row represents an hour. 
\section{\uppercase{Proposed Approach}}
This section describes the implementation of metaheuristic algorithms from Section 2 for time series forecasting using various models. Each model is trained on a training set to find optimal hyperparameters, evaluated by mean average percentage error (MAPE) on the test data.In preprocessing, missing data was handled. The precipitation amount column had significant missing data and was dropped. The temperature column had a small amount of missing data (0.03\% of total data). Models forecasted 24 hours into the future based on every 3 hours of data. The StandardScaler function from sklearn.pre-processing library \cite{pedregosa2011scikit} was used for standardization. The dataset was split into train dataset ($D_{train}$), validation dataset ($D_{val}$), and test dataset ($D_{test}$). Train data was from Jan 1, 2010, to Dec 31, 2015; validation data from Jan 1, 2016, to Dec 31, 2016; and test data from Jan 1, 2017, to Dec 31, 2020.

\subsection{Auto-Regressive Integrated Moving Average}

We employed the ARIMA $(p,d,q)$ model from statsmodels.tsa.arima.model, utilizing temperature as the sole feature due to its univariate nature. Mean squared error (MSE) served as the fitness function for metaheuristic algorithms such as GA, DE, \& PSO. The hyperparameter search space for each algorithm was limited to $(0,5)$ for $p$, $(0,3)$ for $d$, and $(0,5)$ for $q$, considering our machine's limitations. Differential Evolution yielded the best MAPE of $2.31$.
\subsection{Artificial Neural Networks, Long Short Term Memory, Gated Recurrent Networks}

To ensure consistency among the deep learning models, we maintained a 3-layer architecture with varying neuron counts for input, GRU/LSTM layers, and output. GRU and LSTM models used 8 features from the previous three timesteps (3 hours) as input, with 36, 64, and 24 neurons for input, GRU/LSTM layers, and output, respectively. The ANN model had 24 input features, with 64, 36, and 24 neurons for input, hidden layer, and output, respectively.
\begin{table}[h]
\caption{Best set of hyperparameters (in the order learning rate, batch size, and epochs, respectively) for ANN, LSTM, and GRU averaged over 5 runs.}
\label{tab:my-table}
\centering
\begin{adjustbox}{width=1.0\linewidth, height = 2.1cm}
\begin{tabular}{|>{\centering\arraybackslash}m{2.5cm}|>{\centering\arraybackslash}m{2.3cm}|>{\centering\arraybackslash}m{1.8cm}|>{\centering\arraybackslash}m{1.8cm}|>{\centering\arraybackslash}m{1.8cm}|}
\hline
Metaheuristics                       & Hyperparameters & ANN    & LSTM   & GRU    \\ \hline
\multirow{3}{*}{GA}   & Learning Rate   & 0.0001 & 0.0001 & 0.0001 \\ \cline{2-5} 
                                     & Batch Size      & 80     & 80     & 80     \\ \cline{2-5} 
                                     & Epochs          & 527    & 860    & 758    \\ \hline
\multirow{3}{*}{DE} & Learning Rate & 0.0005 & 0.75   & 0.075  \\ \cline{2-5} 
                                     & Batch Size      & 200    & 24     & 20     \\ \cline{2-5} 
                                     & Epochs          & 8      & 1000   & 200    \\ \hline
\multirow{3}{*}{PSO}     & Learning Rate   & 0.1061 & 0.2964 & 0.4176 \\ \cline{2-5} 
                                     & Batch Size      & 41     & 66     & 202    \\ \cline{2-5} 
                                     & Epochs          & 84     & 33     & 61     \\ \hline
\end{tabular}
\end{adjustbox}
\end{table}

For the ANN model, the dataset shapes were: $x_{train}$: (52558, 24), $y_{train}$: (52558, 24), $x_{val}$: (8760, 24), $y_{val}$: (8760, 24), $x_{test}$: (35016, 24), and $y_{test}$: (35016, 24). For GRU and LSTM models, the dataset shapes considered time steps: $x_{train}$: (53558, 3, 8) and similar shapes for other splits. All networks utilized the ReLU activation function \& employed implemented metaheuristic algorithms to optimize batch size, epochs, and learning rates using Mean Squared Error as the loss metric. The best hyperparameter sets for each model are summarized in Table 1. MAPE plots were generated for each set of hyperparameters.
% \begin{table}[h]
% \begin{adjustbox}{width=8cm, height=1cm, center}
% \begin{tabular}{|
% >{\columncolor[HTML]{FFFFFF}}c 
% >{\columncolor[HTML]{FFFFFF}}c 
% >{\columncolor[HTML]{FFFFFF}}c 
% >{\columncolor[HTML]{FFFFFF}}c |}
% \hline
% \multicolumn{4}{|c|}{\cellcolor[HTML]{FFFFFF} HYPERPARAMETERS FOR ANN, LSTM, GRU}                         \\ \hline
% Metaheuristics & ANN & LSTM               & GRU        \\ \hline
% Genetic Algorithm (GA) & (0.0001, 80, 527) & (0.0001, 80, 860) & (0.0001, 80, 758) \\ \hline
% Differential Evolution (DE)   & (0.0005, 200, 8)  & (0.75, 24, 1000)  & (0.075, 20, 200)  \\ \hline
% Particle Swarm (PSO)   & (0.1061 , 41, 84)  & (0.2964, 66, 33)  & (0.4176, 202, 61) \\ \hline
% \end{tabular}
% \end{adjustbox}
% \caption{Best set of hyperparameters for ANN, LSTM and GRU averaged over 5 tests}
% \label{tab:my-table}
% \end{table}
\section{\uppercase{RESULTS AND DISCUSSION}}

\begin{figure*}[t]
\centering
\includegraphics[height=9 cm,width=15cm]{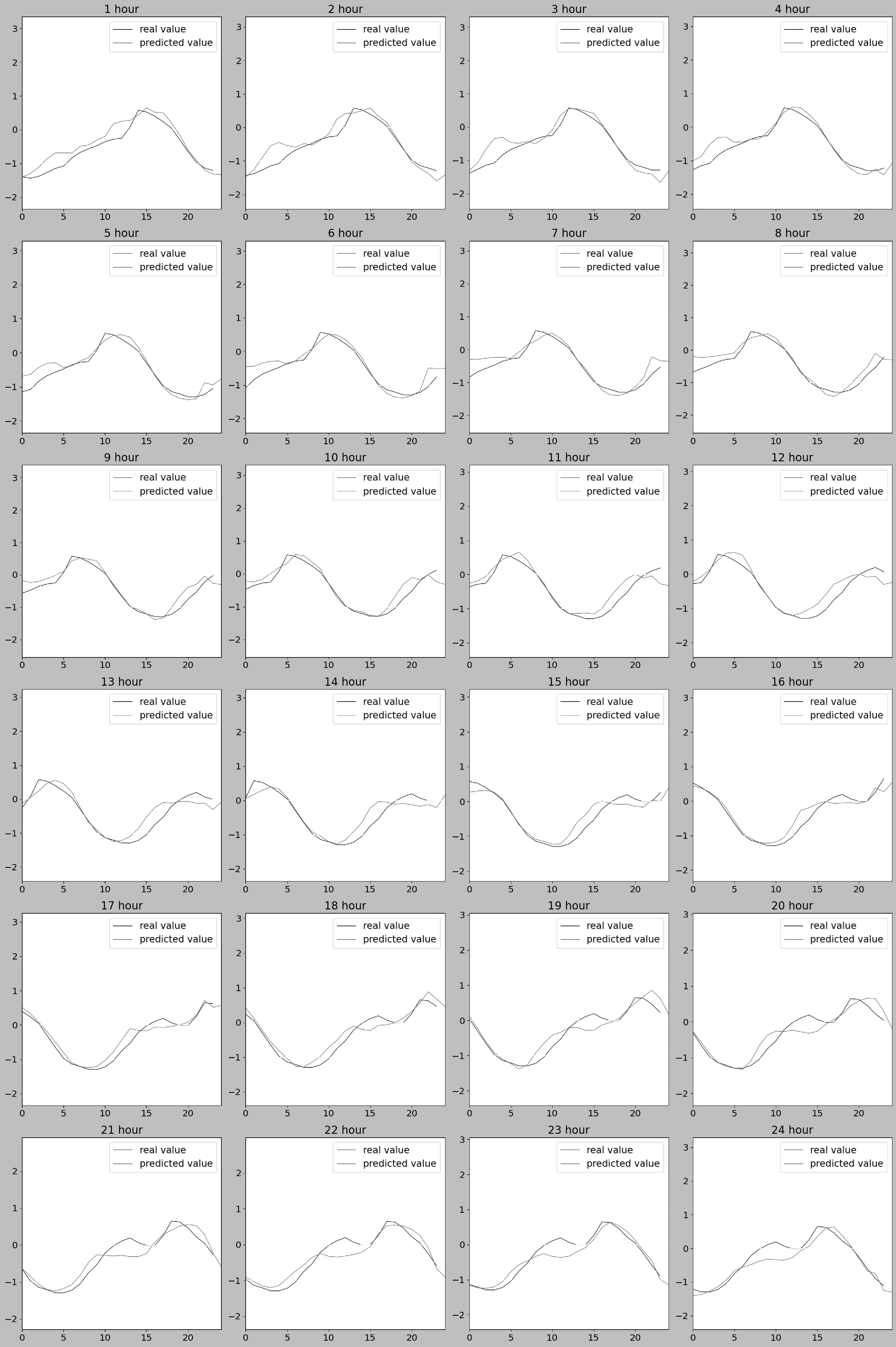}
\caption{Predicted plots for temperature for the next 24 hours starting from the $N$-th hour for the best ANN DE Model.}
\label{Fig-1}
\end{figure*}
 
We implemented the proposed metaheuristics-based optimal hyperparameters' selection, evaluating three deep learning models (ANN, GRU, LSTM) for algorithm suitability while maintaining consistent three-layer architecture. All values shown were averaged over five trials, and all figures displayed used standardized units. Additionally, we explored other models, including the machine learning model ARIMA. The paper's code can be found below.\footnote{Code link: \url{https://github.com/pathikritsyam/ECTA}}

\begin{table}[h!]
\caption{MAPE comparison of different meta-heuristics for ANN, ARIMA, GRU, and LSTM averaged over 5 runs.}
\begin{adjustbox}{width=1.0\linewidth, height = 1.5cm}
\begin{tabular}{|c|l|l|l|l|}
\hline
\multicolumn{5}{|c|}{MAPE - Mean Absolute Percentage Error} \\ \hline
Meta-heuristics & ANN  & ARIMA & GRU  & LSTM \\ \hline
Differential Evolution & 1.15 & 2.31  & 1.75 & 1.65 \\ \hline
Particle Swarm (PSO) & 1.95 & 2.85  & 1.86 & 1.98 \\ \hline
Genetic Algorithm (GA) & 1.97 & 3.28  & 1.93 & 1.97 \\ \hline
Manual Selection & 2.09 & 4.34  & 1.98 & 1.99 \\ \hline
\end{tabular}
\end{adjustbox}
\end{table}

We utilized Standard Scaler for improved convergence and stability during model training with seasonal data. Scaling prevents dominant features and enhances the models' learning effectiveness. Data preprocessing, including scaling and handling seasonal patterns, is crucial for boosting forecasting model performance. The deep learning model's input and output layers have different neuron counts based on requirements. GRU and LSTM use eight features from previous time steps, including the current one, leading to 36, 64, and 24 neurons in the input, neural network, and output layers, respectively. For the ANN model, features from three-hour time steps are concatenated, resulting in an input feature size of 24. Consequently, the input layer has 64 neurons, while the hidden and output layers contain 36 and 24 neurons, respectively. Mean Squared Error (MSE) is employed to minimize training loss. The experiments show that the ANN model optimized with Differential Evolution (DE) outperforms other models with different optimization algorithms. DE with ANN achieves the lowest Mean Absolute Percentage Error (MAPE) of 1.15, followed by DE with LSTM. DE consistently outperforms GA and PSO in terms of MAPE for each model, and performs superiorly across all models compared to GA and PSO.
\begin{figure}[htbp]
  \centering
  \includegraphics[width=7.4cm, height=3.62cm]{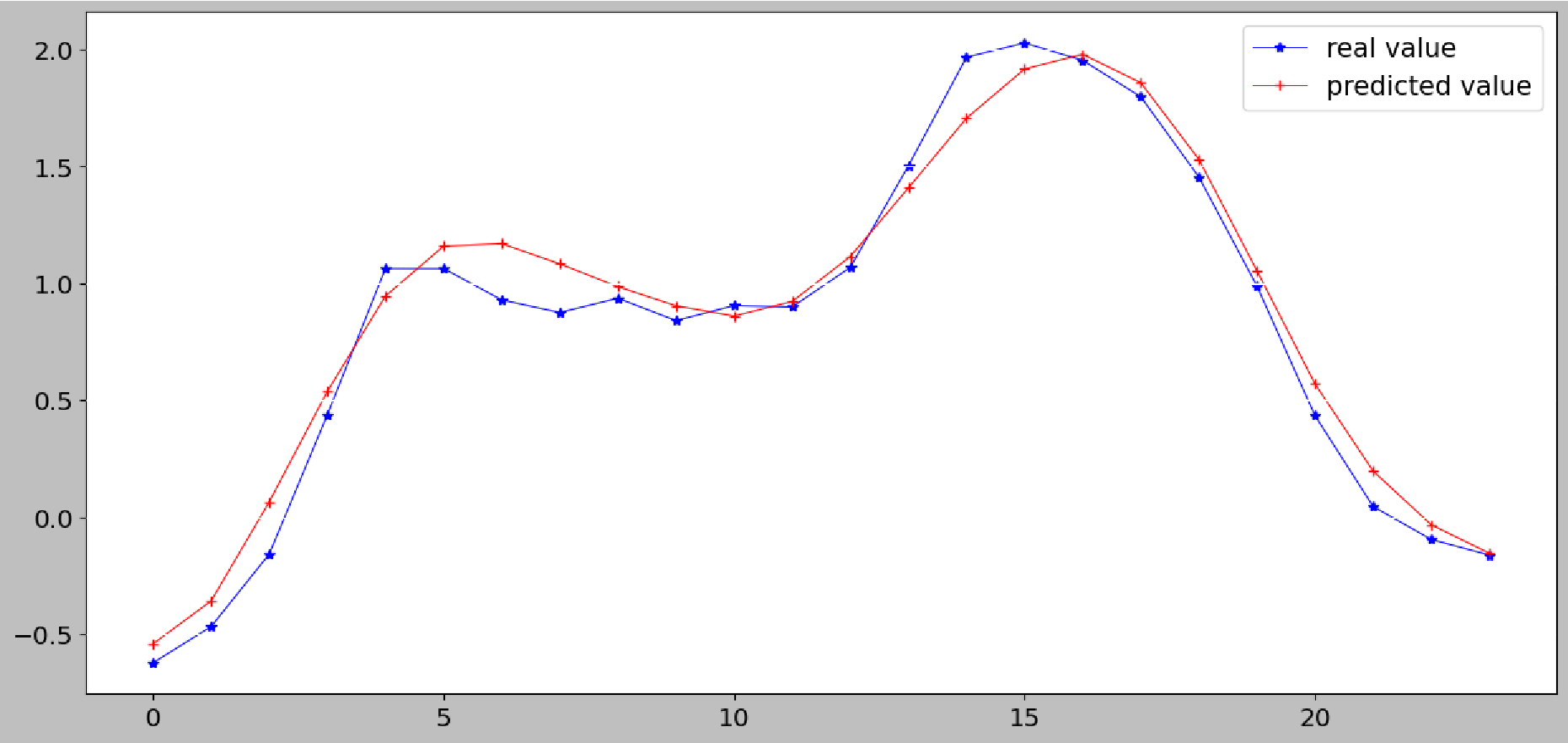}
  \caption{24-hour ahead forecast plot for the best performing model}
  \label{fig:2} % Place the label after the caption
\end{figure}
DE is chosen as the effective metaheuristic algorithm due to its efficient search space exploration and adaptability across generations. Its capability to handle continuous parameter spaces makes it well-suited for optimizing neural network hyperparameters. 
Nevertheless, the optimal optimization algorithm choice may vary depending on the dataset and task. The metaheuristic algorithms consistently outperform manual hyperparameter search. Additionally, deep learning models (ANN, GRU, LSTM) outperform ARIMA in forecasting accuracy. ARIMA's limitation is its inability to capture complex non-linear patterns in data, resulting in inferior performance. The results clearly show that Differential Evolution (DE) outperforms both Particle Swarm Optimization (PSO) and Genetic Algorithm (GA) in terms of Mean Absolute Percentage Error (MAPE) across different models used in the study. DE's superior performance can be attributed to several key factors. It effectively explores the search space and exploits promising regions for optimal solutions. The mutation operator introduces random perturbations to prevent early convergence. The crossover operator facilitates the exchange of promising features, speeding up the convergence process. The selection operator preserves the fittest individuals, enhancing the quality of solutions. PSO demonstrates good performance but falls slightly behind DE. It suffers from premature convergence, limiting its ability to reach the global optimum, which could be an explanation for its performance. GA, on the other hand, shows relatively poor performance compared to both DE and PSO. It could possibly be due to slow convergence, and the fixed encoding scheme may limit its ability to effectively search through the vast hyperparameter space if the solution requires a specific combination of hyperparameters. The findings indicate that deep learning models (ANN, GRU, LSTM) outperform the ARIMA model in forecasting accuracy.

\begin{figure}[htbp]
  \centering
  \includegraphics[width=7.4cm, height=4.2cm]{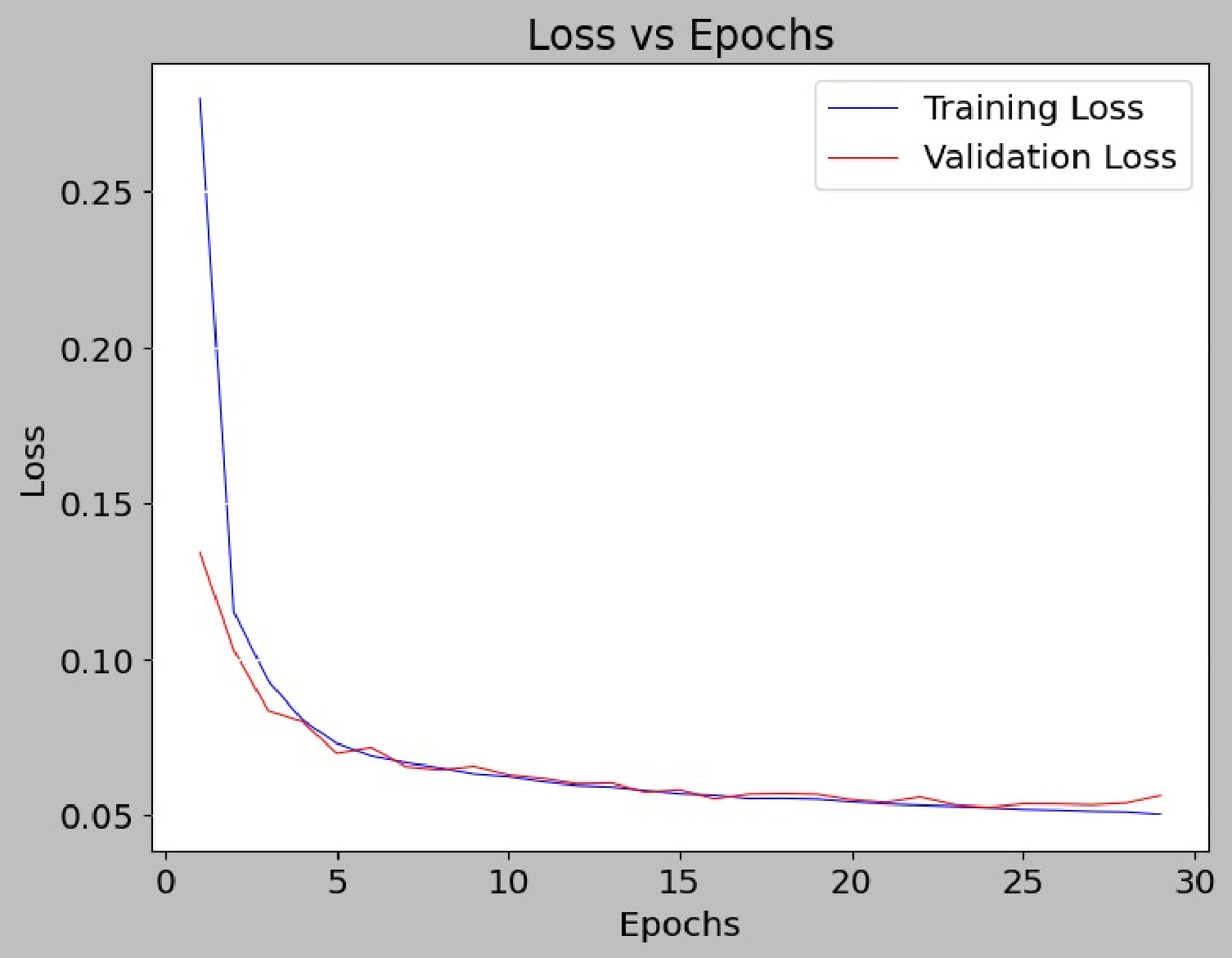}
  \caption{Training \& Validation loss plots vs Epochs for the Best ANN DE Model}
  \label{fig:3} % Place the label after the caption
\end{figure}

Metaheuristic algorithms (DE, GA, PSO) consistently outperform manual hyperparameter search. Among the three, DE proves to be the most effective algorithm, outperforming both GA and PSO.

\section{\uppercase{Conclusion}}
\label{sec:conclusion}
This paper applies metaheuristic algorithms to optimize hyperparameters in deep learning models like Artificial Neural Networks, GRUs, LSTMs, and ARIMA for better performance. We find that Differential Evolution (DE) outperforms Genetic Algorithm (GA) and Particle Swarm Optimization (PSO) in short-term weather forecasting. DE's ability to explore and exploit the search space effectively leads to optimal solutions. While PSO performs well, it can suffer from premature convergence, and GA may have slow convergence and limitations for hyperparameter configurations. In the future, this approach can be extended to explore other evolutionary-based feature selections for various time series applications.
\bibliographystyle{apalike}
{\small
\bibliography{example}}

\begin{thebibliography}{}

\bibitem[Brockwell and Davis, 1996]{Brockwell_Davis_1996}
Brockwell, P.~J. and Davis, R.~A. (1996).
\newblock Arma models.
\newblock {\em Introduction to Time Series and Forecasting}, page 81–108.

\bibitem[Bäck and Schwefel, 1993]{Bäck_Schwefel_1993}
Bäck, T. and Schwefel, H.-P. (1993).
\newblock An overview of evolutionary algorithms for parameter optimization.
\newblock {\em Evolutionary Computation}, 1(1):1–23.

\bibitem[Canada and Change, 2023]{Canada_2023}
Canada, E. and Change, C. (2023).
\newblock Government of canada / gouvernement du canada.

\bibitem[Harvey, 1990]{Harvey_1990}
Harvey, A.~C. (1990).
\newblock Arima models.
\newblock {\em Time Series and Statistics}, page 22–24.

\bibitem[Hochreiter and Schmidhuber, 1997]{Hochreiter_Schmidhuber_1997}
Hochreiter, S. and Schmidhuber, J. (1997).
\newblock Long short-term memory.
\newblock {\em Neural Computation}, 9(8):1735–1780.

\bibitem[Jing et~al.,
  2019]{Jing_Gulcehre_Peurifoy_Shen_Tegmark_Soljacic_Bengio_2019}
Jing, L., Gulcehre, C., Peurifoy, J., Shen, Y., Tegmark, M., Soljacic, M., and
  Bengio, Y. (2019).
\newblock Gated orthogonal recurrent units: On learning to forget.
\newblock {\em Neural Computation}, 31(4):765–783.

\bibitem[Katoch et~al., 2020]{Katoch_Chauhan_Kumar_2020}
Katoch, S., Chauhan, S.~S., and Kumar, V. (2020).
\newblock A review on genetic algorithm: Past, present, and future.
\newblock {\em Multimedia Tools and Applications}, 80(5):8091–8126.

\bibitem[Kelley, 1960]{Kelley_1960}
Kelley, H.~J. (1960).
\newblock Gradient theory of optimal flight paths.
\newblock {\em ARS Journal}, 30(10):947–954.

\bibitem[Kennedy and Eberhart, 1995]{Kennedy_Eberhart_1995}
Kennedy, J. and Eberhart, R. (1995).
\newblock Particle swarm optimization.
\newblock {\em Proceedings of ICNN’95 - International Conference on Neural
  Networks}, 4:1942–1948.

\bibitem[Maclaurin et~al., 2015]{maclaurin2015gradientbased}
Maclaurin, D., Duvenaud, D., and Adams, R.~P. (2015).
\newblock Gradient-based hyperparameter optimization through reversible
  learning.

\bibitem[Man et~al., 1996]{Man_Tang_Kwong_1996}
Man, K., Tang, K., and Kwong, S. (1996).
\newblock Genetic algorithms: Concepts and applications [in engineering
  design].
\newblock {\em IEEE Transactions on Industrial Electronics}, 43(5):519–534.

\bibitem[Masum et~al.,
  2021]{Masum_Shahriar_Haddad_Faruk_Valero_Khan_Rahman_Adnan_Cuzzocrea_Wu_2021}
Masum, M., Shahriar, H., Haddad, H., Faruk, M.~J., Valero, M., Khan, M.~A.,
  Rahman, M.~A., Adnan, M.~I., Cuzzocrea, A., and Wu, F. (2021).
\newblock Bayesian hyperparameter optimization for deep neural network-based
  network intrusion detection.
\newblock {\em 2021 IEEE International Conference on Big Data (Big Data)}.

\bibitem[Nematzadeh et~al., 2022]{NEMATZADEH2022107619}
Nematzadeh, S., Kiani, F., Torkamanian-Afshar, M., and Aydin, N. (2022).
\newblock Tuning hyperparameters of machine learning algorithms and deep neural
  networks using metaheuristics: A bioinformatics study on biomedical and
  biological cases.
\newblock {\em Computational Biology and Chemistry}, 97:107619.

\bibitem[Orive et~al., 2014]{article}
Orive, D., Sorrosal, G., Borges, C., Martin, C., and Alonso-Vicario, A. (2014).
\newblock Evolutionary algorithms for hyperparameter tuning on neural networks
  models.
\newblock {\em 26th European Modeling and Simulation Symposium, EMSS 2014},
  pages 402--409.

\bibitem[Pedregosa et~al., 2011]{pedregosa2011scikit}
Pedregosa, F., Varoquaux, G., Gramfort, A., Michel, V., Thirion, B., Grisel,
  O., Blondel, M., Prettenhofer, P., Weiss, R., Dubourg, V., et~al. (2011).
\newblock Scikit-learn: Machine learning in python.
\newblock {\em Journal of machine learning research}, 12(Oct):2825--2830.

\bibitem[Radzi et~al., 2021]{Radzi_Karim_Saripan_Rahman_Isa_Ibahim_2021}
Radzi, S.~F., Karim, M.~K., Saripan, M.~I., Rahman, M.~A., Isa, I.~N., and
  Ibahim, M.~J. (2021).
\newblock Hyperparameter tuning and pipeline optimization via grid search
  method and tree-based automl in breast cancer prediction.
\newblock {\em Journal of Personalized Medicine}, 11(10):978.

\bibitem[Shekar and Dagnew, 2019]{8882943}
Shekar, B.~H. and Dagnew, G. (2019).
\newblock Grid search-based hyperparameter tuning and classification of
  microarray cancer data.
\newblock In {\em 2019 Second International Conference on Advanced
  Computational and Communication Paradigms (ICACCP)}, pages 1--8.

\bibitem[Storn and Price, 1997]{Storn_Price_1997}
Storn, R. and Price, K. (1997).
\newblock Differential evolution – a simple and efficient heuristic for
  global optimization over continuous spaces - journal of global optimization.

\bibitem[Wang et~al., 2022]{Wang_Ma_Xu_Wang_Wu_2022}
Wang, S., Ma, C., Xu, Y., Wang, J., and Wu, W. (2022).
\newblock A hyperparameter optimization algorithm for the lstm temperature
  prediction model in data center.
\newblock {\em Scientific Programming}, 2022:1–13.

\end{thebibliography}

\end{document}